# SOLVING NMF WITH SMOOTHNESS AND SPARSITY CONSTRAINTS USING PALM

## A PREPRINT

**Raimon Fabregat**
Laboratoire de Physique
École normale supérieure de Lyon
46, allée d'Italie 69364 Lyon cedex 07 France
raimon.fa@gmail.com

**Pierre Borgnat**
Laboratoire de Physique
École normale supérieure de Lyon
46, allée d'Italie 69364 Lyon cedex 07 France
pierre.borgnat@ens-lyon.fr

**Paulo Gonçalves**
Laboratoire De L'Informatique Du Parallelisme
École normale supérieure de Lyon
46, allée d'Italie 69364 Lyon cedex 07 France
pierre.borgnat@ens-lyon.fr

**Nelly Pustelnik**
Laboratoire de Physique
École normale supérieure de Lyon
46, allée d'Italie 69364 Lyon cedex 07 France
nelly.pustelnik@ens-lyon.fr

March 18, 2021

## ABSTRACT

Non-negative matrix factorization is a problem of dimensionality reduction and source separation of data that has been widely used in many fields since it was studied in depth in 1999 by Lee and Seung , including in compression of data , document clustering , processing of audio spectrograms and astronomy. In this work we have adapted a minimization scheme for convex functions with non-differentiable constraints called PALM to solve the NMF problem with solutions that can be smooth and/or sparse, two properties frequently desired.

*K*eywords NMF · Sparsity · Smoothness · PALM

## 1 Introduction

Non-negative Matrix Factorization (NMF) is an algorithm that decompose objects as a combination of common smaller pieces, or images as combination of different patterns. Mathematically, NMF consists on approximating a non-negative matrix $\mathbf{V} \in \mathbb{R}^{D \times N}$ as the product of two other non-negative matrices $\mathbf{W} \in \mathbb{R}^{D \times K}$ and $\mathbf{H} \in \mathbb{R}^{K \times N}$. The parameter $K$ is defined a priori, and is usually set such that $DK + DN << FN$. Intuitively, the problem consists on trying to express a set of N samples $\mathbf{v}_i \in \mathbb{R}^D$, the columns of $\mathbf{V}$, as a linear combinations of K non-negative "meta" vectors $\mathbf{w}_i \in \mathbb{R}^D$, the columns of $\mathbf{W}$, where the coefficients of the linear combination are in the columns of $\mathbf{H}$. That is:

$$\mathbf{V} \approx \mathbf{W}\mathbf{H} \quad \text{and} \quad \mathbf{v}_i \approx \sum_{j=1}^{K} \mathbf{w}_j \mathbf{H}_{ji} \tag{1}$$

The non-negativity of the decomposition is useful in the cases where non-negativity is inherent in the original data, such as in image processing and document classification, where negative values are not meaningful. In such cases, the factorization decomposes the elements into arch-types that are directly interpretable: a usual application in image processing is to use the algorithm to find $K$ sub-images or patterns (the columns of $\mathbf{W}$) a linear combination of which can approximately reconstruct a collection of N images (the columns of $\mathbf{V}$).

Finding this decomposition can be done by solving the following optimization problem

$$\operatorname*{argmin}_{\mathbf{W}\in\mathbb{R}^{D\times K},\mathbf{H}\in\mathbb{R}^{K\times N}} d(\mathbf{V}|\mathbf{WH})$$
$$\text{subject to } 0 \preceq \mathbf{W}, 0 \preceq \mathbf{H}$$

(2)

where $d(\mathbf{V}|\mathbf{WH})$ is an objective function that is minimum and 0 when $\mathbf{V} = \mathbf{WH}$.

Different objective functions have been proposed to assess the factorization depending on the type of noise that $\mathbf{V}$ is supposed to be contaminated with. The most popular cost functions used for the NMF are the ones belonging to the $\beta$-divergence family, which include the Euclidean distance ($\beta = 2$), the Kullback-Leibler divergence ($\beta = 1$), and the Itakura-Saito divergence ($\beta = 0$). As noted by Févotte and Cemgil [1], the values $\beta = 2, 1, 0$, are suited for the cases of Gaussian additive, Poisson, and multiplicative gamma noise respectively. In this work we have used the more common $\beta = 2$, although all the derivations can be made for the other cases.

NMF algorithms suffers from rotational indeterminacy, as pointed out in [1], which means that the best solution may not be unique, as one can see:

$$\mathbf{WH} = (\mathbf{WR})(\mathbf{R}^{-1}\mathbf{H}) = \mathbf{W}'\mathbf{H}'$$

The matrices $\mathbf{W}'$ and $\mathbf{H}'$ are restricted to be non-negative though, which significantly reduce the number of possible solutions. The case where both $\mathbf{R}$ and $\mathbf{R}^{-1}$ are non-negative is not a concern, as it is easy to prove that in that case $\mathbf{R}$ is restricted to be a generalized permutation matrix, so that the only thing that $\mathbf{R}$ would change would be the order and scale of the columns of $\mathbf{W}$ and the rows of $\mathbf{H}$ (which is basically the same result). The only problem appears if there exist a combination $\mathbf{R}$ and $\mathbf{R}^{-1}$ without sign restrictions such that $\mathbf{W}'$ and $\mathbf{H}'$ are still non-negative. This possibility is usually ignored as in general the chances of such a matrix to exist are small, and not a concern in face of the bigger issues that the problem has for not being convex, issues that we discuss in the following paragraph, but can be addressed by adding extra constraints to the objective function that would make the degeneration of minimizers disappear[2].

The cost functions of NMF are not convex, which means that if the factorization is solved using convex optimization techniques, the final solution will depend on the initial value used for the optimization process. Results of other factorizations the Singular Value Decomposition as input point can be used to control and accelerate the optimization process[3, 4]. Alternatively, extra constraints can be added to guide the algorithm towards a result with particular characteristics.

Many algorithms to solve the NMF have been developed since the introduction of the problem in 1999 by Lee and Seung[5]. Variations of the original problem have been introduced in order to add different characteristics such as sparsity to the obtained solutions, although specific minimization shcemes are often derived for each specific added constraint.

In this work we have adapted the cost function of NMF in order to promote different degrees of smoothness and sparsity in the results of the factorization. To do so, we have added extra terms to the standard cost function of NMF, and followed the work of Bolte *et al.*[6] to and derive an adequate minimization scheme. Bolte's minimization scheme Proximal Alternating Linearized Minimization (PALM) is a solid framework for non-convex and non-smooth multivariate optimization problems and allows to minimize a cost function with an arbitrary number of smooth and non-smooth constraints, ensuring that the process will converge. The procedure to derive the optimization steps in PALM is systematic and opens the door to combine even more constraints adding additional properties to the obtained factorization.

## 1.1 Smoothness and sparsity constraints

**Sparse patterns** A usual desired attribute in decompositions like NMF is that the patterns in which the signals are decomposed are as sparse as possible. Sparsity can improve the interpretability of the patterns learned, and can reduce improve compression tasks. Sparsity can be promoted by adding a term proportional to the $L_1$ norm of $\mathbf{W}$ to the function to minimize. Adding the $L_1$ in this way mimics the $L_0$ norm (i.e. the number of elements in the variable that are non-zero) and promotes sparsity[7].

**Smooth activation coefficients** Often, two adjacent columns of $\mathbf{V}$ are adjacent in time. In such cases, it may be desirable to limit the rate of change between the representation of two adjacent elements of $\mathbf{V}$, in case it is known that the evolution of the signals cannot be too fast. This can be done using the Tikhonov regularization, meaning adding an extra term proportional to $\|\mathbf{H}\boldsymbol{\Gamma}\|_F^2$ to the objective function, where $\boldsymbol{\Gamma}$ is the following matrix:

$$\boldsymbol{\Gamma} = \begin{bmatrix} 1 & 0 & \dots & 0 \\ -1 & \ddots & \ddots & \vdots \\ 0 & \ddots & \ddots & 0 \\ \vdots & \ddots & \ddots & 1 \\ 0 & \dots & 0 & -1 \end{bmatrix}$$

Basically this extra term "encourage" the algorithm to minimize the difference between the coefficients of two adjacent elements.

With these constraints we need to add also extra terms to control that the values inside the matrices $\mathbf{W}$ and $\mathbf{H}$ do not become too large. This is necessary as, for example when using the sparsity variant, the algorithm could make the $L_1$ norm of $\mathbf{W}$ arbitrarily small just scaling down all the values in the matrix $\mathbf{W}$ and scaling up $\mathbf{H}$ by the same factor; on the other side, with the smoothness constraint, the algorithm could arbitrarily reduce $\|\mathbf{W}\boldsymbol{\Gamma}\|_F^2$ scaling down $\mathbf{H}$ and scaling up $\mathbf{W}$. Controlling the values in the matrices can easily be done just adding terms proportional to $\|\mathbf{W}\|_F^2$ and $\|\mathbf{H}\|_F^2$.

So the function to optimize becomes

$$\Psi(\mathbf{W}, \mathbf{H}) = \|\mathbf{V} - \mathbf{W}\mathbf{H}\|_F^2 + \eta\|\mathbf{H}\boldsymbol{\Gamma}\|_F^2 + \delta_{\mathbf{W}\succeq 0} + \lambda\|\mathbf{W}\|_1 + \delta_{\mathbf{H}\succeq 0} + \beta_{\mathbf{W}}\|\mathbf{W}\|_F^2 + \beta_{\mathbf{H}}\|\mathbf{H}\|_F^2 \qquad (3)$$

Minimizing this cost function is not trivial, as the $L_1$ norm is not a smooth function. To do so, we have relied on PALM and derived an adequate minimization scheme.

## 1.2 PALM

The following is a short description of the PALM framework and its benefits based on Bolte's work. This method is aimed to solve problems of the following form:

$$\underset{x,y}{\text{minimize}}\ \Psi(x, y) := f(x) + g(y) + H(x, y) \qquad (4)$$

where the functions $f$ and $g$ are extended valued (i.e., allowing the inclusion of constraints) and H is a smooth function. The method is not restricted to two variables though, and the results and the properties of the method hold true for any finite number of them.

PALM was developed as a framework to include non-convex and non-smooth settings. It is based on the Proximal Forward-Backward (PFB) algorithm for minimization problems of one variable with non-convex and non-smooth objective functions and constraints. PFB relies on the assumption that the function to be minimized satisfies the Kurdyka-Łojasiewicz property, which allows to derive the convergence of the bounded trajectories of the steepest descent equation to critical points[6]. Basically, satisfying the KL property in a point $a$ means that there exists some $\theta \in [\frac{1}{2}, 1)$ such that the function $|f - f(a)|^\theta\|\nabla f\|^{-1}$ remains bounded around $a$[8], and a *KL function* is a function that satisfies the KL property at every point. The class of functions satisfying such property is very large, and covers a considerable amount of non-convex/non-smooth functions arising in many fundamental applications, including of course all the proper and smooth functions. A more detailed description of the property and the functions satisfying them can be found elsewhere[8].

PFB minimizes the sum of a smooth (differentiable) function $h$ with a non-smooth (non-differentiable) one $\sigma$ of one variable, and consists of a gradient step on the smooth part followed by by a proximal step on the non-smooth part:

$$x^{k+1} \in \operatorname{argmin}_{x\in\mathbb{R}^d}\left\{ <x - x^k, \nabla h(x^k) > + \frac{t}{2}\|x - x^k\|_2^2 + \sigma(x)\right\}, (t > 0),$$

or using the proximal map notation,

$$x^{k+1} \in \operatorname{prox}_t^\sigma(x^k - \frac{1}{t}\nabla h(x^k)) \qquad (5)$$

PALM combines the alternating minimization scheme with the PFB, applying PFB steps alternating between variables. What follows is the pseudo-code of the algorithm as presented by Bolte[6] :

---

**PALM: Proximal Alternating Linearized Minimization**

1. Initialization: start with any $(x^0, y^0)$

2. For each $k = 0, 1, ...$ generate a sequence $\left\{ (x^k, y^k) \right\}_k$ as follows:

   (a) Take $\gamma_1 > 1$, set $c_k = \gamma_1 L(y^k)$ and compute

   $$x^{k+1} \in \text{prox}_{c_k}^f (x^k - \frac{1}{c_k} \nabla_x H(x^k, y^k))$$

   (b) Take $\gamma_2 > 1$, set $d_k = \gamma_2 L(x^{k+1})$ and compute

   $$y^{k+1} \in \text{prox}_{d_k}^g (y^k - \frac{1}{d_k} \nabla_y H(x^{k+1}, y^k))$$

---

where $L(x^k)$ is the Lipschitz constant of $\nabla_x H(x^k, y^k)$, and ensures convergence. The values $\gamma_1$ and $\gamma_2$ are only restricted to be bigger than one. We have used $\gamma_1 = \gamma_2 = 1.1$, but other values can be assigned to them, affecting only the speed of convergence.

This algorithm is proven to converge by Bolte[6], and its benefits are that it gives a solid model that can be applied to a wide variety of problems and its variants, for example to the NMF problem that concerns us.

## 1.3  PALM-NMF

As Bolte already pointed out[6], the original version of the NMF can be easily solved using PALM, for instance taking

$$\Psi(\mathbf{W}, \mathbf{H}) = \|\mathbf{V} - \mathbf{WH}\|_F^2 + \delta_{\mathbf{W} \succeq 0} + \delta_{\mathbf{H} \succeq 0}$$
$$= H(\mathbf{W}, \mathbf{H}) + f(\mathbf{W}) + g(\mathbf{H})$$

where $\delta_{\mathbf{X} \succeq 0}$ is the indicator function of the set $\{\mathbf{X}, \text{ such that } \mathbf{X} \succeq 0\}$, defined as

$$\delta_{\mathbf{X} \succeq 0} = \left\{ \begin{array}{l} 0 \text{ if } \mathbf{X} \succeq 0 \\ \infty \text{ else} \end{array} \right.$$

The variables $x$ and $y$ have been substituted by $\mathbf{W}$ and $\mathbf{H}$ (not to be confused with the function $H$). We have used this notation trying to follow at the same time the one usually used in the NMF literature and the one used in the article by Bolte[6] about PALM.

The PALM ingredients are easily derived:

- $H(\mathbf{W}, \mathbf{H}) = \|\mathbf{V} - \mathbf{WH}\|_F^2 = Tr((\mathbf{V} - \mathbf{WH})(\mathbf{V} - \mathbf{WH})^T)$
  - $\nabla_{\mathbf{W}} H(\mathbf{W}, \mathbf{H}) = 2\mathbf{WHH}^T - 2\mathbf{VH}^T$
  - $\nabla_{\mathbf{H}} H(\mathbf{W}, \mathbf{H}) = 2\mathbf{W}^T \mathbf{WH} - 2\mathbf{W}^T \mathbf{V}$
  - $L(\mathbf{W}) = 2\|\mathbf{HH}^T\|_F$ and $L(\mathbf{H}) = 2\|\mathbf{W}^T \mathbf{W}\|_F$

- $\text{prox}_t^{\delta_{\mathbf{H} \succeq 0}}(\mathbf{H}) = \text{argmin}_{\mathbf{X}} \, \delta_{\mathbf{H} \succeq 0} + \frac{t}{2}\|\mathbf{X} - \mathbf{H}\|_F^2 = P_+(\mathbf{H}) = max\{0, \mathbf{H}\}$ which is nothing more than the projection of H to the $\mathbb{R}_+^{K \times N}$ space, or in even simpler words, setting all the negative values of $\mathbf{H}$ to zero. The same applies for $\mathbf{W}$.

Introducing the previous results to the PALM scheme directly gives the full minimization algorithm. The stopping criteria can either be the relative size of the step being small enough or a fixed number of iterations.

## 1.4  PALM-NMF with smoothness and sparsity

This framework allows also to easily add extra constraints to the objective function to get solutions with different characteristics. As discussed before, the cost function we want to minimize is

$$\Psi(\mathbf{W}, \mathbf{H}) = \|\mathbf{V} - \mathbf{WH}\|_F^2 + \eta\|\mathbf{H\Gamma}\|_F^2 + \delta_{\mathbf{W} \succeq 0} + \lambda\|\mathbf{W}\|_1 + \delta_{\mathbf{H} \succeq 0} + \beta_{\mathbf{W}}\|\mathbf{W}\|_F^2 + \beta_{\mathbf{H}}\|\mathbf{H}\|_F^2 \quad (6)$$

.

The Tikhonov regularization is a smooth function and can be included in the gradient step. On the other side, the $L_1$ norm is not smooth, but can be included in the function $f$, and in the proximal operator. We have used $\beta_{\mathbf{H}} = \beta_{\mathbf{W}} = 0.1$, but any positive value can be assigned to them. The changes for the PALM steps are

- $H(\mathbf{W}, \mathbf{H}) = \|\mathbf{V} - \mathbf{WH}\|_F^2 + \eta\|\mathbf{H\Gamma}\|_F^2 + \beta_{\mathbf{W}}\|\mathbf{W}\|_F^2 + \beta_{\mathbf{H}}\|\mathbf{H}\|_F^2$
    - $\nabla_{\mathbf{H}} H(\mathbf{W}, \mathbf{H}) = 2\mathbf{W}^T\mathbf{WH} - 2\mathbf{W}^T\mathbf{V} + 2\eta\mathbf{H\Gamma\Gamma}^T + 2\beta_{\mathbf{H}}\mathbf{H}$
    - $\nabla_{\mathbf{W}} H(\mathbf{W}, \mathbf{H}) = 2\mathbf{WHH}^T - 2\mathbf{VH}^T + 2\beta_{\mathbf{W}}\mathbf{W}$
    - $L(\mathbf{W}) = 2\|\mathbf{HH}^T\|_F + 2\beta_{\mathbf{H}}$ and $L(\mathbf{H}) = 2\|\mathbf{W}^T\mathbf{W}\|_F + 2\beta_{\mathbf{W}} + 2\eta\|\mathbf{\Gamma\Gamma}^T\|_F$.
- Now $f(\cdot) = \delta_{\succeq 0} + \lambda\|\cdot\|_1$, and it can easily be shown that $\text{prox}_t^f(\cdot) = max\left\{0, \cdot - \frac{2\lambda}{t}\right\}$

## 2 Results

In this section we show the performance of the algorithm and how the added constraint allow to improve the quality of the solution.

### 2.1 Sparsity

To illustrate how the sparsity constraint alters the solution of the NMF we use the algorithm for one of its popular application, the deconstruction of images of faces[9]. To do it we take a set of 2-D pictures of faces, flatten each one of them to a vector, and concatenate them together to form a matrix $\mathbf{V}$ that we can use as input for the NMF. In figure 1a are plotted the pictures of the images of faces that we have used for the decomposition, their approximation after applying the NMF and the patterns learned, for the cases with and without sparsity. The decomposition without sparsity in this case is not very interesting, as we have used a high value of $K$ (the dimensionality is not reduced that much) and the patterns learned are some kind of fusion of faces. In the case with sparsity however, the patterns learned are localized and characterize different parts of the faces, highlighting structures such as the forehead, the hair or the cheekbones, giving an approximation which is almost as good than the one without sparsity, but with patterns that much more interpretable.

### 2.2 Smoothness

Being able to add a smooth evolution of the NMF representation can be beneficial in many different frameworks. As an example, we consider a toy scenario where a priory it is known that the coefficients of the linear combination, the rows of $\mathbf{H}$, are smooth functions. We create a non-negative matrix $\mathbf{V}$ using a random matrix $\mathbf{W}_r \in \mathbb{R}^{100 \times 5}$ and a matrix $\mathbf{H}_r \in \mathbb{R}^{5 \times 200}$ whose rows ,plotted in figure 2a, are smooth functions. We add also Gaussian noise: $\mathbf{V} = max(\mathbf{W}_r \cdot \mathbf{H}_r + \mathcal{N}(0, \sigma^2), 0)$, setting the negative values that the noise may create to zero to ensure that there are no negative values in $\mathbf{V}$. Then we use the PALM-NMF with the objective function 3 with a positive $\eta$ and no sparsity constraints ($\lambda = 0$). The value of the parameter $\eta$ is arbitrary, the higher the smoother the solutions will be, although, a priori, the "degree" of smoothness of the solution cannot be fixed.

In the figure 2 we can see the benefits of our approach. The kernels learned with the smooth NMF resemble much more the original ones, up to some scaling, even though the approximated $\mathbf{V}$ from the PALM-NMF without smoothness constraints was closer to the real $\mathbf{V}$ than the one obtained with smoothness constraints (the Frobenius norm of the difference between the real matrix $\mathbf{V}$ and the approximation $\mathbf{WH}$, $\|\mathbf{V} - \mathbf{WH}\|_F$ was 53.85 for the NMF with smoothness, and 55.38 without smoothness, so the approximation of the original NMF was better).

### 2.3 Simultaneous smoothness and sparsity

To see that both constraints can be combined, we take again the same toy example as in the smoothness test, building a matrix $\mathbf{V} = |\mathbf{W}_r \cdot \mathbf{H}_r + \mathcal{N}(0, \sigma^2)$ with the rows of $\mathbf{H}_r$ being the smooth functions plotted in figure 2a, but now with $\mathbf{W}_r$ being a sparse matrix, initialized with random values and with 80% of its entries set to zero. Therefore, we want to recover the matrices $\mathbf{W}_r$ and $\mathbf{H}_r$ knowing that the rows of $\mathbf{H}_r$ are smooth and that the matrix $\mathbf{W}_r$ is sparse. To measure the quality of the recovery, we can take the learned matrices $\mathbf{W}$ and $\mathbf{H}$ and compare them to the original ones $\mathbf{W}_r$ and $\mathbf{H}_r$. To do so, we normalize the columns of $\mathbf{W}$ and $\mathbf{W}_r$ and the rows of $\mathbf{W}$ and $\mathbf{H}_r$ and order the columns and rows of $\mathbf{W}$ and $\mathbf{H}$ such that the learned patterns that match the original ones are in the same order in the matrices. In figure 3 is shown the "quality" of the learned normalized and ordered $\hat{\mathbf{W}}$s and $\hat{\mathbf{H}}$s in terms of their distance to the normalized original ones, $\hat{\mathbf{W}}_r$ and $\hat{\mathbf{H}}_r$, using the original NMF without constraints, with the sparsity constraint, with the smoothness constraint and with both of them at the same time. Each variation of the algorithm was ran 15 times with different random initializations. We can see that using both sparsity and smoothness gives the best performance. The constraints not only improve the quality of the recovery but also greatly reduce the variability of the solutions due to the random initializations.

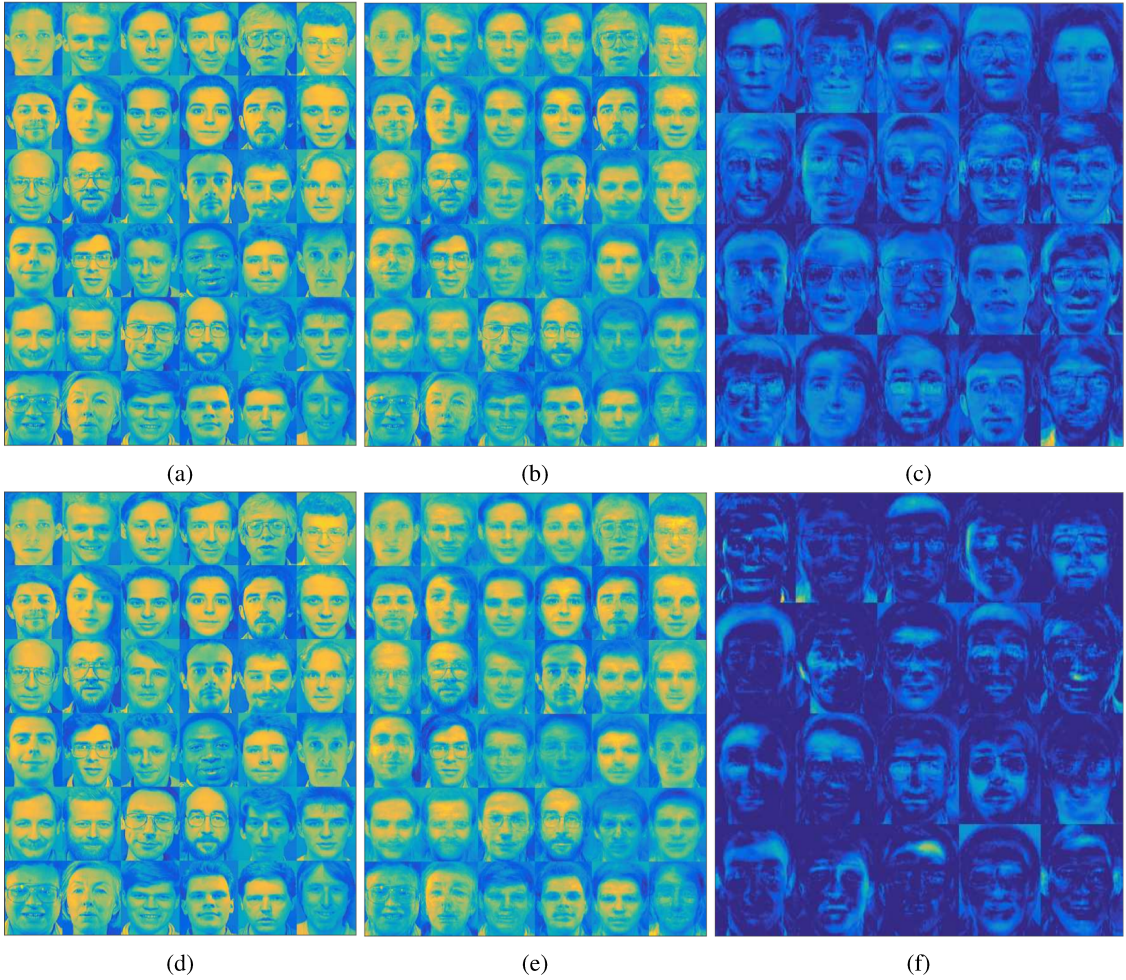

Figure 1: In the figures *a* and *d* the original faces are plotted. Figures *b* and *e* are the faces that the algorithm approximates as a combination of 20 sub-images. Figures *c* and *f* are the sub-images or patterns that the algorithm learned for the cases without sparsity (*c*) and with sparsity constraints (*f*). We see that the sparse patterns characterize different parts of a face, and are able to reconstruct the original faces equally well, to the point that it is hard to notice the differences between *b* and *e*.

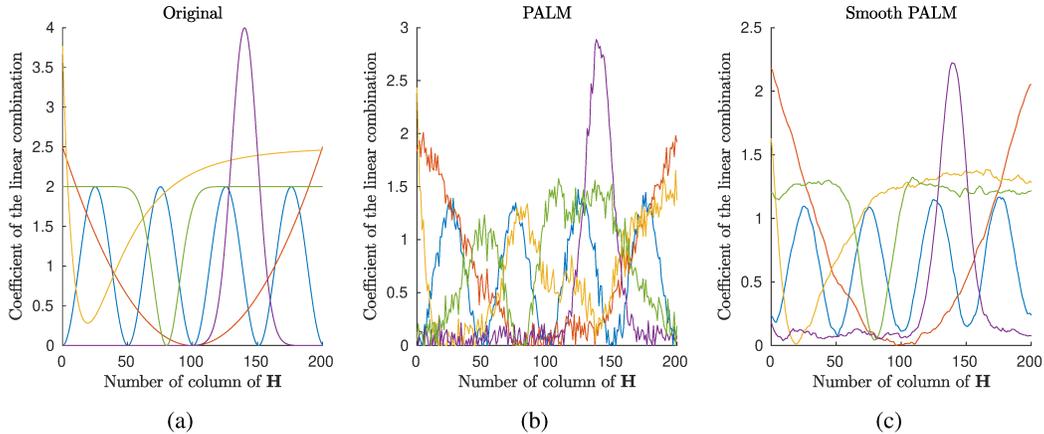

Figure 2: A) original smooth functions stored in the rows of $\mathbf{H}_r$. B) the learned rows of $\mathbf{H}$ with the original PALM-NMF without extra constraints. C) result using the PALM-NMF with the smoothness constraint.

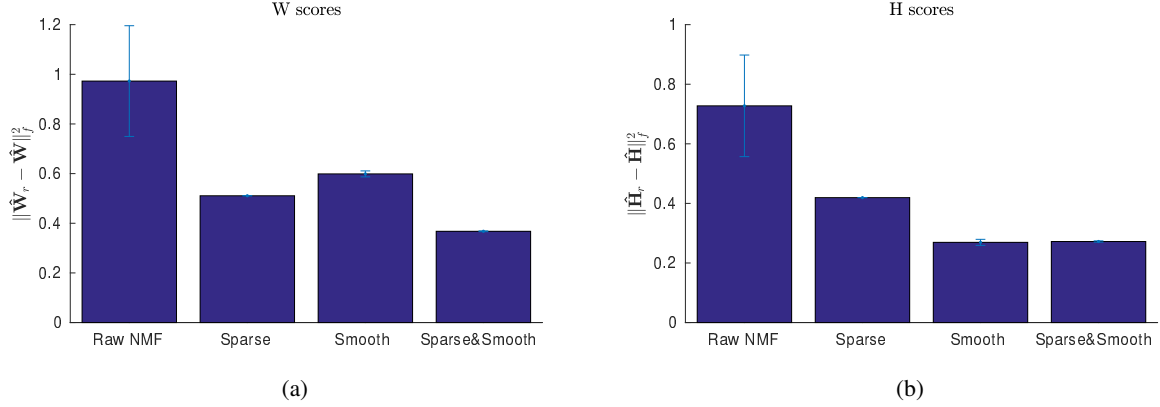

Figure 3: Bar plots with the scores of the different variations of the algorithm on the data generated artificially

## 3    Conclusions

We have adapted the highly versatile optimization scheme PALM to solve the NMF problem with solutions that can be sparse and smooth. While there are other algorithms to solve the NMF with sparsity, it is the first time to the knowledge of the authors that smoothness is added to the NMF solutions. Furthermore, the PALM scheme allows to add multiple constraints at the same time and easily derive the optimization steps.

The code of the algorithm PALM-NMF with the constraints as we presented it can be found in https://github.com/raimonfa/palm-nmf.